\documentclass[conference]{IEEEtran}

\IEEEoverridecommandlockouts
\usepackage{cite}
\usepackage[utf8]{inputenc}
\usepackage{amsmath,amssymb,amsfonts}
\usepackage{algorithmic}
\usepackage{graphicx}
\usepackage{textcomp}
\usepackage{xcolor}
\usepackage{natbib}
\usepackage{booktabs}
\usepackage{tikz}
\usepackage{pgfplots}
\usepackage{hyperref}
\usepackage[symbol]{footmisc}

\definecolor{mygreen}{HTML}{167dde}
\definecolor{myred}{HTML}{f22835}
\colorlet{greenfill}{mygreen!20!white}
\colorlet{redfill}{myred!20!white}
\colorlet{moreredfill}{myred!40!white}
\usetikzlibrary{positioning}
\usetikzlibrary{backgrounds}
\usetikzlibrary{calc}
\usetikzlibrary{fit}

\usepackage{makecell}

\def\BibTeX{{\rm B\kern-.05em{\sc i\kern-.025em b}\kern-.08em
    T\kern-.1667em\lower.7ex\hbox{E}\kern-.125emX}}
\begin{document}

\title{KuroNet: Pre-Modern Japanese Kuzushiji Character Recognition with Deep Learning\\
}

\author{
\IEEEauthorblockN{Tarin Clanuwat*}
\IEEEauthorblockA{\textit{Center for Open Data in the Humanities} \\
\textit{National Institute of Informatics}\\
Tokyo, Japan \\
tarin@nii.ac.jp}
\and
\IEEEauthorblockN{Alex Lamb*}
\IEEEauthorblockA{\textit{MILA} \\
\textit{Université de Montréal}\\
Montreal, Canada \\
lambalex@iro.umontreal.ca
}
\and
\IEEEauthorblockN{Asanobu Kitamoto}
\IEEEauthorblockA{\textit{Center for Open Data in the Humanities} \\
\textit{National Institute of Informatics}\\
Tokyo, Japan \\
kitamoto@nii.ac.jp
}
}

\maketitle

\begin{abstract}
Kuzushiji, a cursive writing style, had been used in Japan for over a thousand years starting from the 8th century.  Over 3 millions books on a diverse array of topics, such as literature, science, mathematics and even cooking are preserved. However, following a change to the Japanese writing system in 1900, Kuzushiji has not been included in regular school curricula. Therefore, most Japanese natives nowadays cannot read books written or printed just 150 years ago.  Museums and libraries have invested a great deal of effort into creating digital copies of these historical documents as a safeguard against fires, earthquakes and tsunamis.  The result has been datasets with hundreds of millions of photographs of historical documents which can only be read by a small number of specially trained experts.  Thus there has been a great deal of interest in using Machine Learning to automatically recognize these historical texts and transcribe them into modern Japanese characters. Nevertheless, several challenges in Kuzushiji recognition have made the performance of existing systems extremely poor.  To tackle these challenges, we propose KuroNet, a new end-to-end model which jointly recognizes an entire page of text by using a residual U-Net architecture which predicts the location and identity of all characters given a page of text (without any pre-processing). This allows the model to handle long range context, large vocabularies, and non-standardized character layouts. We demonstrate that our system is able to successfully recognize a large fraction of pre-modern Japanese documents, but also explore areas where our system is limited and suggest directions for future work.  
\end{abstract}


\section{Introduction}
Kuzushiji or cursive style Japanese characters were used in the Japanese writing and printing system for over a thousand years. However, the standardization of Japanese language textbooks (known as the  \textit{Elementary School Order}) in 1900 \citep{takahiro2013meiji}, unified the writing type of Hiragana and also made the Kuzushiji writing style obsolete and incompatible with modern printing systems.  Therefore, most Japanese natives cannot read books written just 150 years ago. 

\footnotetext{*Equal Contribution} 

However, according to the General Catalog of National Books \citep{shoten1963catalog} there are over 1.7 million books written or published in Japan before 1867. Overall it has been estimated that there are over 3 million books preserved nationwide \citep{clanuwat2018classical}.  The total number of documents is even larger when one considers non-book historical records, such as personal diaries. Despite ongoing efforts to create digital copies of these documents, most of the knowledge, history, and culture contained within these texts remain inaccessible to the general public. One book can take years to transcribe into modern Japanese characters. Even for researchers who are educated in reading Kuzushiji, the need to look up information (such as rare words) while transcribing as well as variations in writing styles can make the process of reading texts time consuming.  Additionally entering the text into a standardized format after transcribing it requires effort.  For these reasons the vast majority of these books and documents have not yet been transcribed into modern Japanese characters.  

\subsection{A Brief Primer on the History of the Japanese Language}

We introduce a small amount of background information on the Japanese language and writing system to make the recognition task more understandable.  Since Chinese characters entered Japan prior to the 8th century, the Japanese wrote their language using Kanji (Chinese characters in the Japanese language) in official records. However, from the late 8th century, the Japanese began to add their own character sets: Hiragana and Katakana, which derive from different ways of simplifying Kanji. Individual Hiragana and Katakana characters don't contain independent semantic meaning, but instead carry phonetic information (like letters in the English alphabet).  The usage of Kanji and Hiragana in the Heian period (794 - 1185 A.D.) were separated by the gender of the writer: Kanji was used by men and Hiragana was used by women.  Only in rare cases such as that of Lady Murasaki Shikibu, the author of \textit{the Tale of Genji}, did women have any Kanji education. This is why stories like \textit{the Tale of Ise} or collections of poems such as \textit{Kokinwakashū} were written in mostly Hiragana and official records were written in Kanbun (a form of Classical Chinese used in Japan written in all Kanji without Hiragana).  This is the main reason why the number of character classes in Japanese books can greatly fluctuate.  Some books have mostly Kanji while other books have mostly Hiragana. Additionally, Katakana was used for annotations in order to differentiate them from the main text.  


The oldest printed document which has survived in Japan is \textit{Hyakumanto Darani}, a Sutra from the 8th century. Since then Japan has used two printing systems: movable types and woodblock prints. Woodblock printing dominated the printing industry in Japan until the 19th century. Even though these books were printed, characters carved in the block were copied from handwritten documents.   Therefore woodblock printed characters are very similar to handwritten ones.  

\begin{figure}
\vskip -0.05in
\begin{center}
\centerline{\includegraphics[width=0.9\columnwidth]{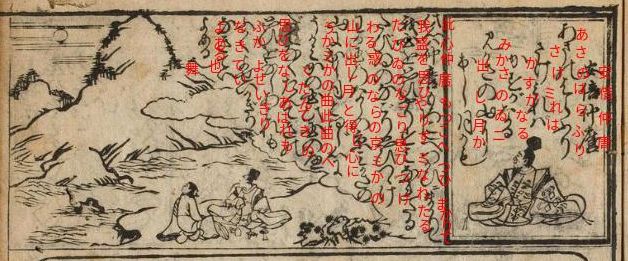}}
\vskip -0.01 in
\caption{KuroNet recognition result of \textit{Isemonogatari Taisei} \citep{isetaisei}. The model achieved an F1-score on this page of 0.8824.  However, some parts are hard to recognize correctly as a result of the book's condition.}
\label{fig:isetaiseiocr}
\end{center}
\vskip -0.25 in
\end{figure}

Since woodblocks were created from a whole piece of wood, it was easy to integrate illustrations with texts.  Thus many books have pages where text is wrapped around illustrations (an example is shown in Figure~\ref{fig:isetaiseiocr}).  

Another note is that the Japanese primarily used a woodblock for printing style.This had some important influences on the types of documents produced.  One is that it made it easy to integrate illustrations or calligraphy into text, and indeed this is relatively common.  Additionally, it had the impact of making each press's copy of a given book visually distinct.  

\begin{figure}[!htb]
\vskip -0.05in
\begin{center}
\centerline{\includegraphics[width=0.9\columnwidth]{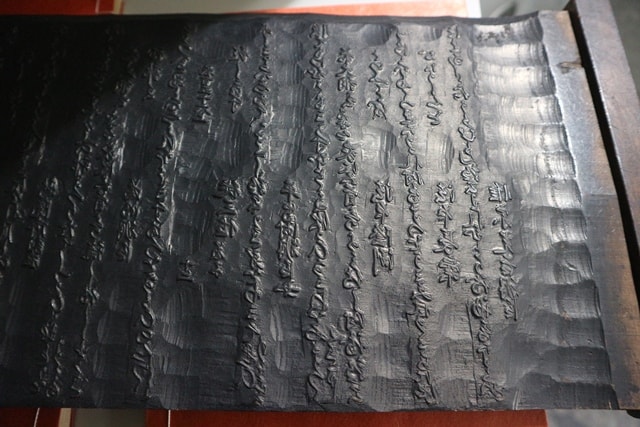}}
\vskip -0.05 in
\caption{An example of a Woodblock for book printing from the 18th century (Hanawa Hokiichi Museum, Tokyo)}
\label{fig:hokiichiwoodblock}
\end{center}
\vskip -0.02 in
\end{figure}

\begin{figure}[!htb]
\vskip -0.25in
\begin{center}
\centerline{\includegraphics[width=0.9\columnwidth]{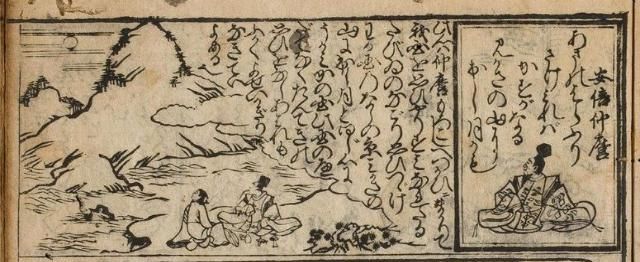}}
\vskip -0.05 in
\caption{Woodblock printed book, \textit{Isemonogatari Taisei} \citep{isetaisei} showing an example of how text in woodblock prints can wrap around illustrations.  }
\label{fig:isetaisei}
\end{center}
\vskip -0.05 in
\end{figure}

Our method offers the following contributions: 

\begin{itemize}
    \item A new and general algorithm for pre-modern Japanese document recognition which uses no pre-processing and is trained using character locations instead of character sequences.  
    \item Demonstration of our novel algorithm on recognizing pre-modern Japanese texts.  Our results dramatically exceed the previous state-of-the-art.  Out of the 27 books evaluated on held-out pages: 
        \begin{itemize}
            \item 15\% had an F1-score between 90\%-100\%
            \item 56\% had an F1-score between 80\%-90\%
            \item 22\% had an F1-score between 70\%-80\%
            \item 7\% had an F1-score between 58\%-70\%
        \end{itemize}  
    \item A solution to recognize Kuzushiji text even when it includes illustrations (an example result is shown in Figure~\ref{fig:isetaiseiocr}).  
    \item An exploration of why our approach is well-suited to the challenges associated with the historical Japanese recognition task.  
\end{itemize}

\section{KuroNet}

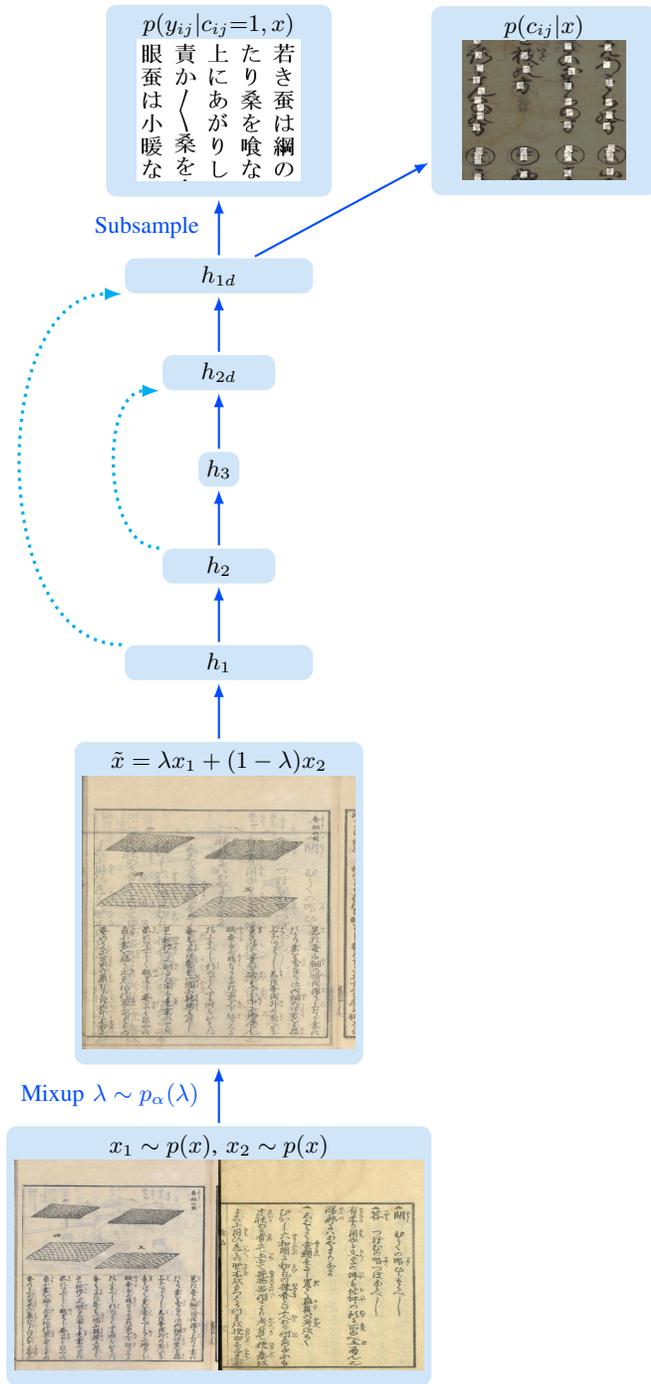
\begin{figure}[ht!]
\centering
\begin{tikzpicture}[
    scale=0.9,
    black!50, text=black,
    font=\small,
    every node/.append style={font=\small},
    node distance=1mm,
    dnode/.style={
        align=center,
        rectangle,minimum size=15mm,rounded corners,
        inner sep=20pt},
    rnode/.style={
        align=center,
        rectangle,
        minimum width=5mm,
        minimum height=4mm,
        rounded corners,
        inner sep=3pt,
        thin, draw=none},
    tuplenode/.style={
        align=center,
        rectangle,minimum size=10mm,rounded corners,
        inner sep=15pt},
    darrow/.style={
        rounded corners,-latex,shorten <=5pt,shorten >=1pt,line width=2mm},
    mega thick/.style={line width=2pt}
    ]
    
\matrix[row sep=8.2mm, column sep=0mm] {
    \node (y) [rnode,left color = greenfill, right color = greenfill, minimum width=30mm] {\makecell[c]{$p(y_{ij} \vert c_{ij}{=}1,x)$\\\includegraphics[width=.12\textwidth,clip]{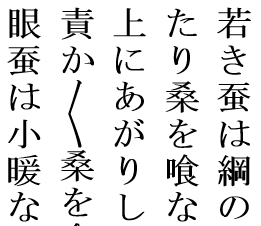}}}; &&
    \node (yprox) [rnode,left color = greenfill, right color = greenfill, minimum width=30mm] {\makecell[c]{$p(c_{ij} \vert x)$\\\includegraphics[width=.12\textwidth,trim={7cm 8.3cm 7cm 7cm},clip]{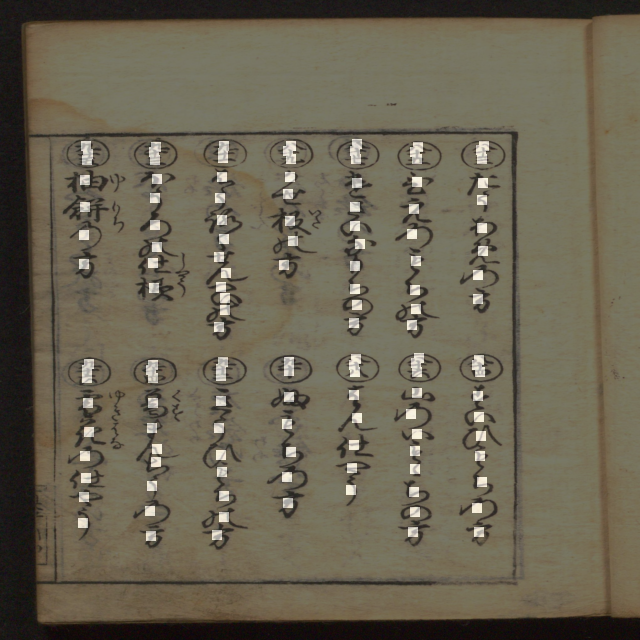}}}; &&
     \\
    \node (h1d) [rnode,left color = greenfill, right color = greenfill, minimum width=25mm] {$h_{1d}$}; &&
     \\
    \node (h2d) [rnode,right color=greenfill, left color = greenfill, minimum width=15mm] {$h_{2d}$};&&
     \\
    \node (h3) [rnode,right color=greenfill, left color = greenfill] {$h_{3}$};&&
     \\
    \node (h2) [rnode,left color = greenfill, right color = greenfill, minimum width=15mm] {$h_2$}; &&
     \\
    \node (h1) [rnode,left color = greenfill, right color = greenfill,minimum width=25mm] {$h_1$};&&
     \\
    \node (xm) [rnode,left color = greenfill, right color = greenfill, minimum width=30mm] {\makecell[c]{$\tilde{x} = \lambda x_1 + (1-\lambda) x_2$\\\includegraphics[width=.2\textwidth,trim={1cm 1cm 1cm 1cm},clip]{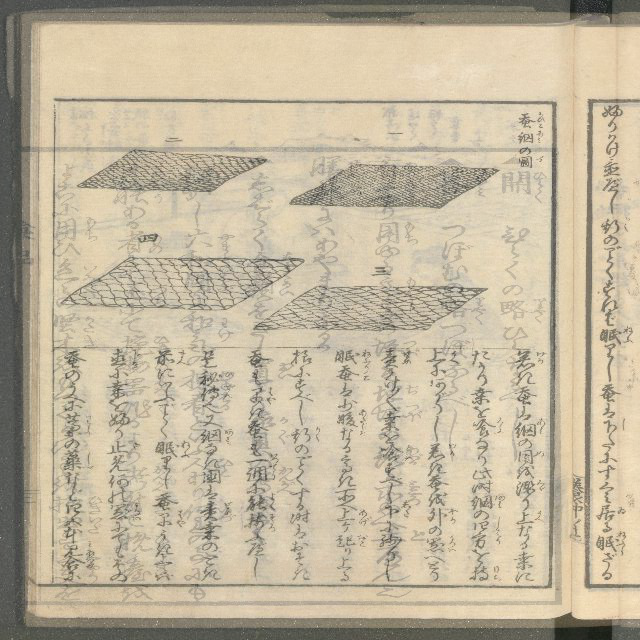}}};&&
      \\
    \node (input) [rnode,left color = greenfill, right color = greenfill, minimum width=30mm] {\makecell[c]{$x_1 \sim p(x)$, $x_2 \sim p(x)$\\\includegraphics[width=.15\textwidth,trim={1cm 1cm 1.5cm 1cm},clip]{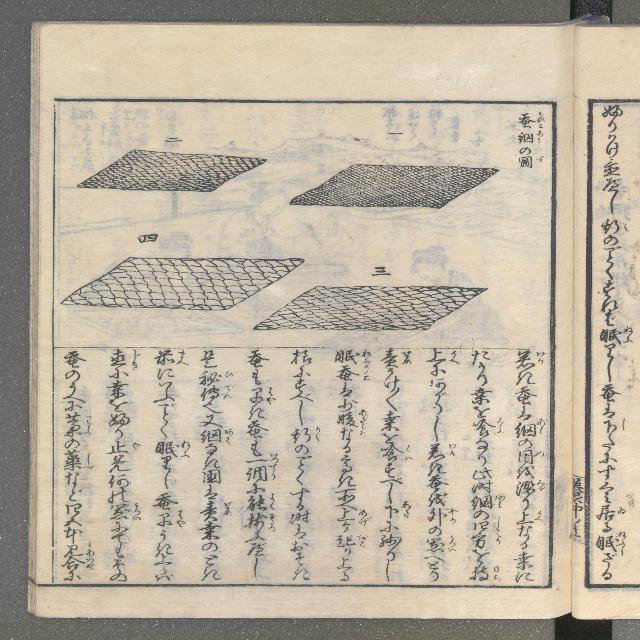}\includegraphics[width=.15\textwidth,trim={0.5cm 1cm 2cm 1cm},clip]{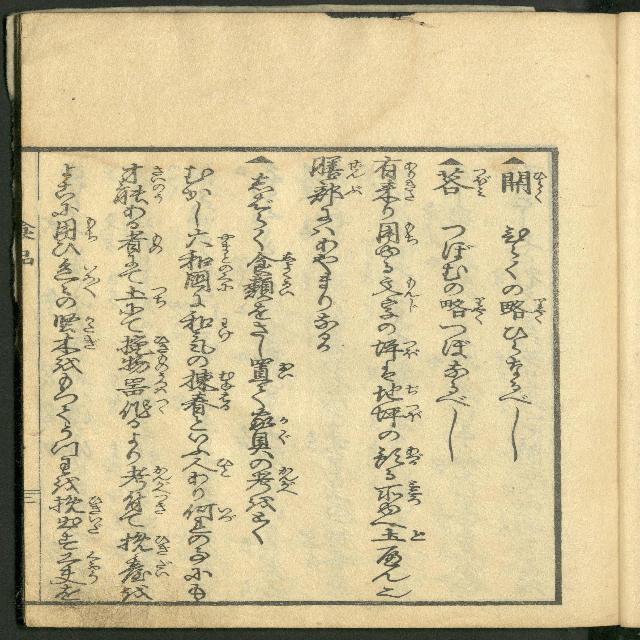}}};&&
     \\
};
\draw[-latex,shorten <=1pt,shorten >=1pt,thick,color=blue!70!cyan]  (input) to node[label=left:Mixup $\lambda \sim p_{\alpha}(\lambda)$]{} (xm);
\draw[-latex,shorten <=1pt,shorten >=1pt,thick,color=blue!70!cyan]  (xm) to (h1);
\draw[-latex,shorten <=1pt,shorten >=1pt,thick,color=blue!70!cyan]  (h1) to (h2);
\draw[-latex,shorten <=1pt,shorten >=1pt,thick,color=blue!70!cyan] (h2) to (h3);
\draw[-latex,shorten <=1pt,shorten >=1pt,thick,color=blue!70!cyan] (h3) to (h2d);
\draw[-latex,shorten <=1pt,shorten >=1pt,thick,color=blue!70!cyan] (h2d) to (h1d);
\draw[-latex,shorten <=1pt,shorten >=1pt,thick,color=blue!70!cyan] (h1d) to node[label=left:Subsample]{} (y);
\draw[-latex,shorten <=1pt,shorten >=1pt,thick,color=blue!70!cyan] (h1d) to (yprox);



\draw[-latex,shorten <=1pt,shorten >=1pt, very thick,color=cyan, dotted] (h1.north west)  to [out=180,in=180]  (h1d.south west);

\draw[-latex,shorten <=1pt,shorten >=1pt, very thick,color=cyan, dotted] (h2.north west)  to [out=180,in=180] (h2d.south west);



\end{tikzpicture}
\caption{The KuroNet algorithm illustrated.  On the left we show how a residual U-Net is used to compute $p(c|x)$ and $p(y|c{=}1,x)$ starting from an image X.  Our algorithm uses skip connections to capture global context and local information, the mixup regularizer, and uses no pre-processing of the image.  }
\end{figure}

The KuroNet method is motivated by the idea of processing an entire page of text together, with the goal of capturing both long-range and local dependencies.  KuroNet first rescales each given image to a standardized size of 640 x 640.  Then the image is passed through a U-Net architecture to obtain a feature representation of size C x 640 x 640, where C is the number of channels in the final hidden layer (in our experiments we used C = 64).  

We refer to the input image as $x \sim p(x)$ and refer to the correct character at each position (i,j) in the input image as $y_{ij} \sim P(y_{ij} \vert x)$.  

We can model each $P(y_{ij} \vert x)$ as a multinomial distribution at each spatial position.  This requires the assumption that characters are independent between positions once we've conditioned on the image of the page.  Another issue is that the total number of characters in our dataset is relatively large (4000), so storing the distribution-parameters of the multinomial at each position is computationally expensive.  For example at a 640x640 resolution this requires 6.5 gigabytes of memory just to store these values.  

To get around this, we introduce an approximation where we first estimate if a spatial position contains a character or if it is a background position.  We can write this Bernoulli distribution as $P(c_{ij} \vert x)$, where $c_{ij}{=}1$ indicates a position with a character and $c_{ij}=0$ indicates a background position.  Then we model $P(y_{ij} \vert c_{ij}{=}1, x)$, which is the character distribution at every position which contains a character.  

This allows us to only compute the relatively expensive character classifier at spatial positions which contain characters, dramatically lowering memory usage and computation.  This approximation is called Teacher Forcing and it has been widely studied in the machine learning literature \citep{lamb2016professor,goyal2017actual}.  The Teacher Forcing algorithm is statistically consistent in the sense that it achieves a correct model if each conditional distribution is correctly estimated, but if the conditional distributions have errors, these errors may compound.  To give a concrete example in our task, if our estimate of $p(c_{ij} \vert x)$ has false positives, then during prediction we will evaluate $p(y_{ij} \vert c_{ij}{=}1, x)$ at positions where no characters are present and which the character classifier was not exposed to during training.  Besides computation, one advantage to using teacher forcing is that simply estimating if a position has a character or not is a much easier task than classifying the exact character, which may make learning easier.  

The result of our training process is two probability distributions: $P(y_{ij} \vert c_{ij}{=}1,x)$ and $P(c_{ij}{=}1 \vert x)$.  As we produce estimates that may be inconsistent across different spatial positions, we use clustering as a post-processing step to ensure that each character image in the text is only assigned a single class.  To do this we used the DBSCAN clustering algorithm \citep{ester1996density}, which does not require the number of classes to be tuned as a hyperparameter, and we report the hard cluster centers.  

\subsection{Training and Architecture Details}

We trained for 80 epochs with a batch size of one.  We found that using higher resolution images improved results qualitatively, so we elected to use larger resolutions even though this required us to use a batch size of 1 to stay within GPU memory.  We used the Adam optimizer \citep{kingma2014adam} with a learning rate of 0.0001, $\beta_1$ of 0.9, and $\beta_2$ of 0.999.  

We used the residual FusionNet variant \citep{quan2016fusionnet} of the U-Net architecture \citep{ronneberger2015u} to compute the features.  We used four downsampling layers followed by four upsampling layers, with skip connections passing local information from the downsampling layers to the upsampling layers.  We used 64 channels in the first downsampling layer, and doubled the number of channels with each downsampling layer, following the procedure used by \citep{ronneberger2015u}.  On each upsampling layer, we halved the number of channels.  Our final hidden representation had 64 channels at each position.  

The most significant change we made to the \cite{ronneberger2015u,quan2016fusionnet} architecture was to replace all instances of batch normalization \citep{ioffe2015batch} with group normalization \citep{wu2018group}.  Batch normalization makes the assumption that every batch has the same batch statistics (that each feature will have the same mean and variance).  This approximation is often poor when the batch size is small, which is especially true in our case as we used a batch size of one.  While our results with batch normalization were okay, we noticed that the model often struggled with unusual pages - for example in pages containing large illustrations it would struggle to not predict characters over those parts of the image.  We found that these issues were resolved by using group normalization.  When using group normalization, we always used 16 groups.  

\subsection{Regularization}
\label{sec:reg}

\begin{figure*}[!htb]
    \centering
    \includegraphics[width=0.32\linewidth]{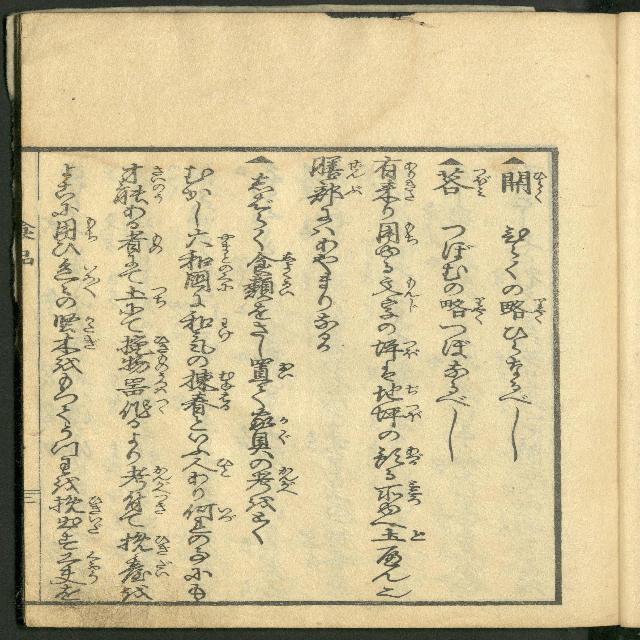}
    \includegraphics[width=0.32\linewidth]{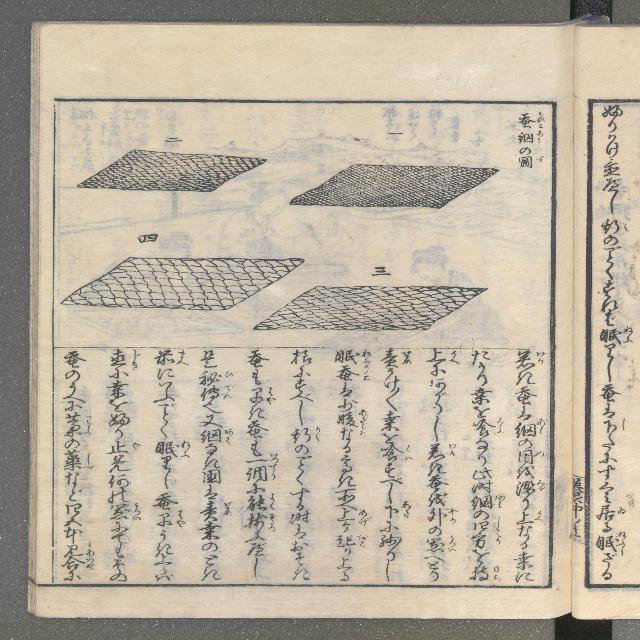}
    \includegraphics[width=0.32\linewidth]{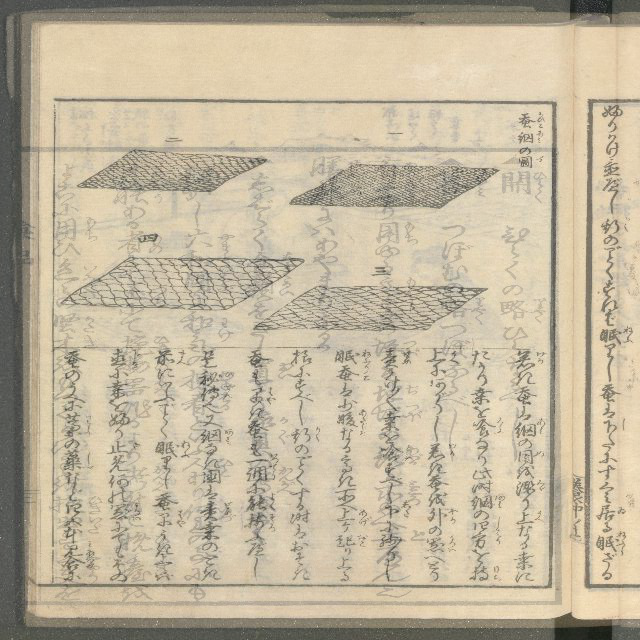}
    \caption{An example of images used during training with the mixup regularizer \citep{zhang2017mixup}.  We mix the left image (30\%) with the middle image (70\%) to produce the right image (mixed).  30\% is the strongest mixup rate that we could have sampled during training.  }
    \label{fig:mixup}
\end{figure*}

We also added a few simple regularizers to improve generalization performance.  First we used mixup to slightly modify the input images \citep{zhang2017mixup}.  We used a variant where the labels are unchanged but the mixing distribution is $Beta(\alpha, \alpha+1)$.  We set $\alpha=0.4$ and to be conservative we clamped the resulting $\alpha$ to be between 0.0 and 0.3.  Our goal with this setup was to encourage the model to only mix in a small amount from different examples while retaining the original label.  Thus we picked relatively conservative hyperparameters that led to images which did not make it too hard to read the original image.  \cite{verma2018manifold} explored the possibility of underfitting when doing Mixup, and we wanted to avoid this possibility.  We provide an example of mixup with $\lambda=0.3$ in Figure~\ref{fig:mixup}.  

Many books are written on a relatively thin paper, so the content of the adjacent page is often faintly visible through the paper.  This can be seen in the center image in Figure~\ref{fig:mixup}.  While it is somewhat subjective, the images produced by mixup appear somewhat similar to images where the adjacent page's content are faintly visible.  Thus Mixup may have an added benefit of helping to encourage the model to ignore the adjacent page, as it is not useful for correct character recognition.  

As an additional regularizer we slightly perturbed the brightness on each update between 90\% and 110\%, to simulate variations in lighting when the books were photographed as well as differences in the darkness of the paper (which are not relevant for the recognition task).

\section{Related Work}

Previous approaches have considered aspects of the kuzushiji recognition task, although no complete end-to-end system has been proposed prior to this work.  

\textbf{Segment and Classify Characters Individually}.  One approach that has been explored is to first segment the image into patches for individual characters, and then classify each patch separately.  This approach is very computationally appealing, but is inappropriate for the general Kuzushiji recognition task due the contextual nature of many characters.  This was explored by \cite{nguyen2017kuzu}.  \cite{clanuwat2018classical} produced datasets consisting of individually segmented kuzushiji characters, but did not consider the Kuzushiji recognition task in general.  

\textbf{Sequence Models}.  A widely studied approach to handwriting recognition involves learning a sequence model over the characters which is conditioned on the image of the text \citep{bezerra2012rnn}.  This was explored for Japanese historical documents specifically by \cite{le2018kuzu}, but using documents between 1870 and 1945 which characters in the their dataset, while old style prints, are not Kuzushiji, but closer to modern Japanese characters.  

The sequence modeling approach has a major limitation for Kuzushiji because the layout of the text is not necessarily sequential, and in many cases trying to produce a sequential ordering for the text would lead to substantial ambiguity.  The dataset that we used did not have an explicit sequential structure (see Section~\ref{sec:data}).  Another disadvantage is that generating characters sequentially usually requires sampling the characters one-by-one and in-order.  This prediction may be slower than prediction with our model, which processes the entire page in parallel.  

An approach specifically for character spotting using U-Nets was proposed in \citep{clanuwat2018jinmoncom}.  However this considered detecting only 10 chosen Hiragana characters and thus was not an attempt to make the document readable, but rather to pick out a few specific common characters.  

\section{Challenges in Kuzushiji Recognition Task}
\label{sec:challenges}

We identify several challenges in Kuzushiji which make it challenging for standard handwriting recognition systems and explain how KuroNet addresses these challenges.  

\textbf{Context: } Kuzushiji characters are written in such a way that some characters can only be recognized by using context, especially the identity of the preceding characters.  For example, a simple line can indicate a few different characters, depending on the preceding characters.  Because KuroNet uses both local and global context, it is able to disambiguate by using the surrounding characters.  This is a challenge for models which segment the characters before classifying them individually.  
    
\textbf{Large Number of Characters: } The total number of characters in Kuzushiji is very large (our dataset contains 4,645 characters), but their distribution is long-tailed and a substantial fraction of the characters only appear once or twice in the dataset.  This is due to the unique structure of the Japanese language - which consisted at the time of two type of character sets, a phonetic alphabet (with simple and extremely common characters) and non-phonetic Kanji characters.  Kanji consists of both common and rare characters: some of them are highly detailed, while others are just one or two straight lines.  
The large number of characters presents challenges for both computation and generalization.  KuroNet addresses this challenge computationally by using Teacher Forcing to only evaluate the character classifier at positions where characters are present.  However, it may still present a challenge for generalization, because many Kanji characters only appear a few times in the training data.  

\textbf{Hentaigana: } One characteristic of Classical Hiragana or \textit{Hentaigana} (the term literally translates to ``Character Variations'') which has a huge effect on the recognition task is that many characters which can be written a single way in modern Japanese could be written in multiple different ways in pre-modern Japanese.  This is one reason why the pre-modern Japanese variant of the MNIST dataset \citep{clanuwat2018classical} is much more challenging than the original MNIST dataset \citep{lecun1998mnist}.  Many of the characters in pre-modern Japanese have multiple ways of being written, so successful models need to be able to capture the multi-modal distribution of each class.  

\textbf{Sayre's Paradox: } A well-studied problem in document recognition \citep{sayre1973paradox} is that with cursive writing systems the segmentation and recognition tasks are entangled.  This is successfully handled by sequence models which use the entire image as context (for example, using attention) or convolutional models which have access to larger context from the page.  This provides further motivation for KuroNet using the same hidden representations to predict $p(y_{ij} \vert c_{ij}, x)$ and $p(c_{ij} \vert x)$.  

\textbf{Annotations vs. Main Text: } Pre-modern Japanese texts, especially in printed books from the Edo period (17th to 19th century), are written such that annotations are placed between the columns of the main text (usually in a smaller font).  These annotations mostly act as a guide for readers on how to pronounce specific Kanji and were written in either Hiragana or Katakana (in our dataset from the Edo Period, Hiragana was more commonly used). For our task we did not consider annotations to be part of the text to be recognized since our dataset doesn't contain labels for annotations (and if it did, we would still want to discriminate between the main text and annotations).  In this setting, our model needs to use context and content to discriminate between the main text and the annotations so that the annotations can be ignored.  

\begin{figure}
\vskip -0.05in
\begin{center}
\centerline{\includegraphics[width=0.9\columnwidth]{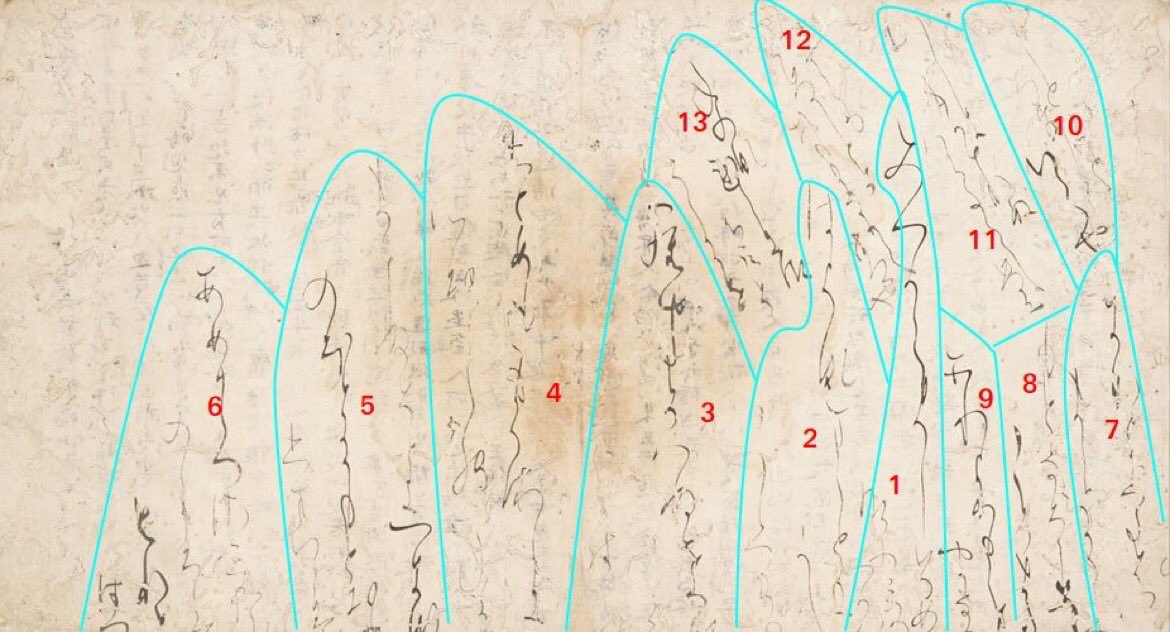}}
\vskip -0.01 in
\caption{An example showing how the text is not necessarily laid out in a clear sequential order.  The red numbers indicate an ordering for how the text should be correctly read.  }
\label{fig:reading_order}
\end{center}
\vskip -0.25 in
\end{figure}

\textbf{Layout:} The layout of Kuzushiji characters does not follow a single simple rule, so it is not always trivial (or even possible) to express the characters as a sequence.  Some examples of this include characters being written to wrap around or even integrate into illustrations.  Another example is a writing style used for personal communications where the ordering of the characters is based on the thickness of their brush strokes in addition to their spatial arrangement.  An example of this is shown in Figure~\ref{fig:reading_order}.  Still another practice involves the use of coded symbols to indicate breaks and continuation in text.  This is a major challenge for systems that assume the data is in a sequence.  As KuroNet does not make this assumption, it does not have any general difficulty with this, and our model performs well on several books which have non-sequential layouts.  Additionally our data (Section~\ref{sec:data}) did not have any sequential information in its labels and in practice we found that using heuristics to estimate the sequential ordering was quite challenging, especially due to the presence of annotations.  

\section{Experiments}

The primary goal of our experiments is to validate the extent to which KuroNet is able to successfully recognize real historical Japanese documents.  We evaluate on two realistic settings: randomly held-out pages from books used during training (Table~\ref{tb:results}) and pages from books never seen by the model during training (Table~\ref{tb:results_overall}).  As secondary goals, we wish to understand the cases where KuroNet doesn't perform well, to motivate future applied research on Kuzushiji, and also to study where it does perform well, to suggest which ideas from KuroNet could be more broadly useful in other areas of document recognition.

\begin{figure*}
    \centering
    \includegraphics[width=0.49\linewidth,trim={0cm 1cm 0cm 2cm},clip]{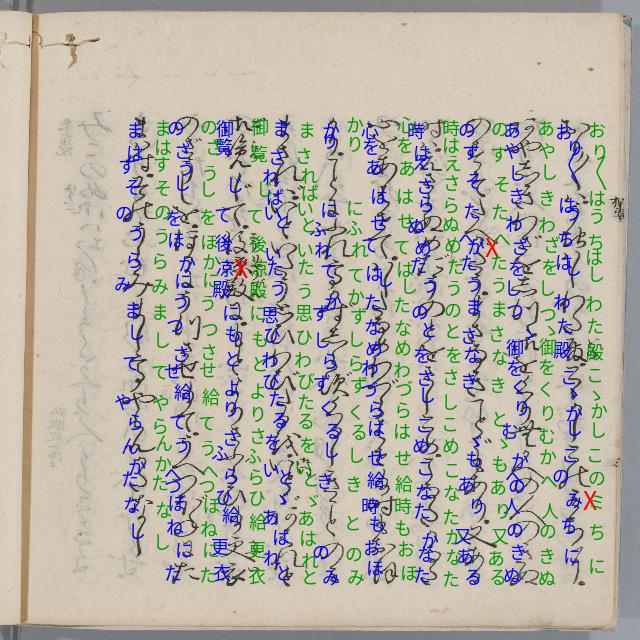}
    \includegraphics[width=0.49\linewidth,trim={0cm 1cm 0cm 2cm},clip]{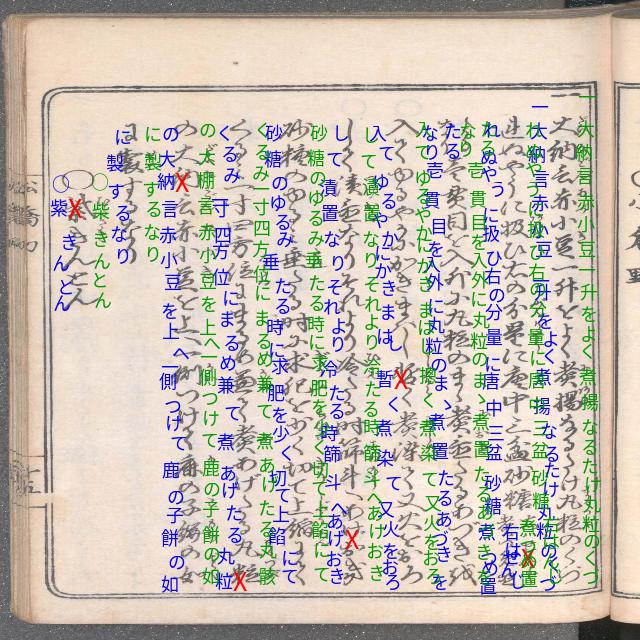} 
    \caption{An example of how our model performs on a page from the test data (held-out pages).  The ground truth is in blue, the model's prediction is in green, and a red X indicates an error.  }
    \label{fig:my_label}
\end{figure*}

\subsection{Data}
\label{sec:data}

The pre-modern Japanese dataset that we use to train and evaluate KuroNet was created by the National Institute of Japanese Literature (NIJL) and released in 2016\footnote{\url{http://codh.rois.ac.jp/char-shape/book/}}.  It is curated by the Center for Open Data in the Humanities (CODH).  The dataset currently has roughly 4,645 character classes and over 684,165 character images.  The dataset consists of bounding boxes for the location of all of the characters in the text as pixel coordinates.  

We also requested to run our model on CODH's private held out set of four books, which were not seen at all by our model during training.  The identity of these books is also not revealed (at present) so that it can provide a fair benchmark for future competitions.  Additionally the number of times that researchers can evaluate on the private dataset is restricted.  

The dataset used in this paper is a precursor to the dataset used for the Kuzushiji Kaggle competition\footnote{https://www.kaggle.com/c/kuzushiji-recognition}, however the datasets are distinct and thus the results reported in this paper are not directly comparable with results on the Kaggle competition.  The Kaggle dataset is somewhat easier than the dataset considered in this paper.  

\subsection{Setup}


We omitted pages which did not have labels in our dataset during training and evaluation.  Generally, pages without labels consisted primarily of an illustration or a cover and a very small amount of text.  We could have included these images in training, and treated the entire page as $c_{ij}={0}$ for training p($c_{ij} \vert x)$.  However as many pages did actually contain a small amount of text, we decided to simply skip the pages out of concern that labeling these points as non-text would lead to too many false negatives in pages containing illustrations.  

In principle KuroNet can train and evaluate on images of different sizes (during both training and evaluation) as it is purely convolutional, but for simplicity we resized every image during training and evaluation to 640 x 640.  Experimentally we found that using larger resolutions substantially improved results.  

\subsection{Quantitative Results}

There are several types of errors that can occur in recognition.  One is that a character is present but our model makes no prediction (false negative).  Another is that the model predicts a character at a position where no character is present or predicts the wrong character when a character is present (false positive).  

We use two basic quantitative metrics to describe our performance: precision and recall.  We define precision as the number of correctly predicted characters divided by the total number of predicted characters.  We define recall as the number of correctly predicted characters divided by the total number of characters present in the ground truth.  We also report the F1-score, which is simply the harmonic mean of the precision and the recall.  

Our dataset contains a non-trivial number of cases where the ground truth dataset is missing some characters which are present.  These may add to the total number of predictions (as our model still predicts for them), which would make our reported precision and recall numbers somewhat worse than they are in reality.  

\begin{figure}[h!]
    \centering
    \includegraphics[width=0.95\linewidth]{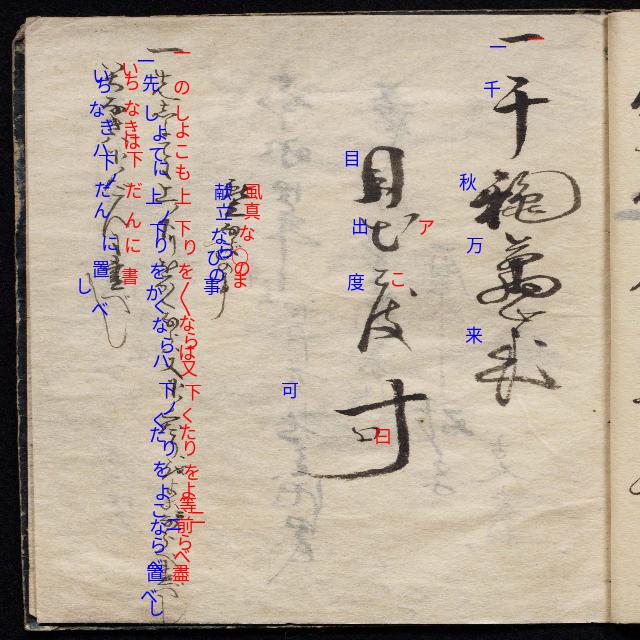}
    \caption{An example of an annotated (ground truth in blue, predictions in red) held-out page from the book where KuroNet performs most poorly (see Table~\ref{tb:results}, Book 200021869).  This cookbook was written in messy handwriting and has many unusually large characters.  }
    \label{fig:200021869l}
\end{figure}

We present our quantitative results in two different tables.  In Table~\ref{tb:results} we show results over the held-out pages (from books used during training), with the results separated out over books.  In Table~\ref{tb:results_overall} we report overall (computed over held-out pages and held-out books) Precision, Recall, and F1-score.  As would be expected, our model's F1-score is somewhat lower on held-out books than on held-out pages.  We also found a considerable variation in performance between books, with over 70\% of of the books having an F1-score over 0.80, but with two books having F1-score below 0.70.  When we investigated which books had the lowest performance, we found that one of the lowest performing books, ``brsk00000'' is a dictionary, and thus contains a large number of rare and obscure words.  This book also has a different layout than other books in that it has many paired columns of text, which our model mistakenly identifies as a single line of text and its annotations (Section~\ref{sec:challenges}).  Another one of the lowest performing books, ``200021869'', is a cookbook written in untidy handwriting. This book also contains rare words and uses unusually large characters when listing recipes (see Figure~\ref{fig:200021869l}).  

For the purposes of understanding which aspects of the model were important for performance, we performed ablation experiments (Table~\ref{tb:results} and Table~\ref{tb:results_overall}).  In the first ablation, we removed mixup.  In the second ablation, we removed mixup and also reduced the model's capacity by using a standard U-Net instead of a residual U-Net \citep{ronneberger2015u} and only predicting the 409 most common characters (instead of the 4000 predicted by KuroNet).  We refer to this second ablation as UNet-Small.  

Intriguingly, we found that adding additional capacity (from the U-Net and higher character count) increased recall substantially but reduced precision on the held-out pages and only slightly improved precision on the held-out books.  Adding the mixup regularizer improved recall and dramatically improved precision on both held-out pages and held-out books, with a larger gain for held-out books.  These results have a simple and satisfying explanation.  Adding additional capacity and predicting more characters improves the model's recall but the model also has poor generalization on rarer characters.  By adding stronger regularization with mixup, the model is able to achieve both better recall and generalize better, leading to improved precision.

\begin{table*}
\centering
\caption{
    Performance of KuroNet across randomly held-out pages from different books.  }
\label{tb:results}
\begin{tabular}{lrrrrrrr} 
\toprule
Book & Book Topic & Year & 
 \shortstack{Number \\of Classes} & \shortstack{Character\\GT}
& \shortstack{UNet-Small\\F1-Score} & \shortstack{ResUNet\\F1-Score} & \shortstack{ResUnet+Mixup\\F1-Score} \\ 
\midrule
100249376 & Food & 1718 & 401 & 1127 & 0.8825 & 0.9184 & \textbf{0.9256} \\
200021644 & Food & 1841 & 785 & 943 & 0.7822 & 0.8885 & \textbf{0.9194}\\
200021712 & Food & 1786 & 843 & 2517 & 0.8270 & 0.8661 & \textbf{0.9120} \\
200021851 & Medical & 1802 & 430 & 429 & 0.8335 & 0.8389 & \textbf{0.9081} \\
100249371 & Food & 1852 & 729 & 972 & 0.7866 & 0.8926 & \textbf{0.8987} \\
hnsd00000 & Literature & 1838 & 1972 & 8518 & 0.8432 & 0.8563 & \textbf{0.8915} \\
100249416 & Food & 1805 & 469 & 1021 & 0.8355 & 0.8659 & \textbf{0.8892} \\
100249537 & Food & 1764 & 826 & 1288 & 0.8089 & 0.8648 & \textbf{0.8857} \\
200014740 & Literature & 1765 & 1969 & 3369 & 0.8144 & 0.8521 & \textbf{0.8849} \\
200003076 & Literature & 1682 & 1720 & 5871 & 0.8092 & 0.8239 & \textbf{0.8792} \\
100241706 & Literature & 1834 & 801 & 1119 & 0.8261 & 0.8315 & \textbf{0.8791} \\
umgy00000 & Literature & 1832 & 1737 & 7746 & 0.8379 & 0.8401 & \textbf{0.8757} \\
200021853 & Food & 1836 & 595 & 839 & 0.8474 & 0.8563 & \textbf{0.8736} \\
100249476 & Food & n/a & 644 & 684 & 0.7818 & 0.8513 & \textbf{0.8713} \\
200015779 & Literature & 1813 & 1817 & 5742 & 0.8291 & 0.8145 & \textbf{0.8631} \\
200021802 & Food & 1643 & 560 & 1981 & 0.8084 & 0.7877 & \textbf{0.8576} \\
200005598 & Literature & 1790 & 660 & 1669 & \textbf{0.8545} & 0.7794 & 0.8235 \\
200004148 & Literature & 1807 & 2061 & 3854 & 0.7582 & 0.7716 & \textbf{0.8136} \\
200022050 & Food & 1684 & 255 & 526 & 0.7897 & 0.7679 & \textbf{0.8122} \\
200021660 & Agriculture & 1803 & 1758 & 3387 & 0.6541 & 0.7143 & \textbf{0.7542} \\
200003967 & Literature & 1878 & 1119 & 1278 & 0.7447 & 0.7051 & \textbf{0.7535} \\
200021925 & Food & 1861 & 693 & 544 & 0.6821 & 0.6481 & \textbf{0.7534} \\
200021763 & Food & n/a & 704 & 1189 & 0.6350 & 0.7015 & \textbf{0.7495} \\
200021637 & Food & n/a & 417 & 422 & 0.6322 & 0.7253 & \textbf{0.7343} \\
200014685 & Literature & 1842 & 1780 & 1287 & 0.6624 & 0.6751 & \textbf{0.7289} \\
brsk00000 & Dictionary & 1775 & 2197 & 7175 & \textbf{0.7436} & 0.6309 & 0.6258 \\
200021869 & Food & n/a & 330 & 216 & \textbf{0.6321} & 0.5471 & 0.5865 \\
\bottomrule
\end{tabular}
\end{table*}

{\renewcommand{\arraystretch}{1.0}
\setlength{\tabcolsep}{5pt}
\begin{table}
\centering
\caption{
    Performance of KuroNet and ablations across different metrics, aggregated over multiple books.  We report results for both held-out pages (from books seen during training) and held-out books (not seen at all during training).  
}
\label{tb:results_overall}
\begin{tabular}{lrrrr} 
\toprule
Book & \shortstack{UNet-Small} & \shortstack{ResUNet} & \shortstack{ResUNet +\\Mixup} \\ 
\midrule
\shortstack{Held-Out Pages\\(473 total pages)}\\
\midrule
Precision &  0.8440 & 0.8182 & \textbf{0.8683} \\
Recall & 0.7901 & 0.8120 & \textbf{0.8417} \\
F1 Score & 0.8162 & 0.8151 & \textbf{0.8548} \\
\midrule
\shortstack{Held-Out Books\\(394 total pages)} \\
\midrule
Precision & 0.7295 & 0.7353 & \textbf{0.8155} \\
Recall & 0.6839 & 0.7239 & \textbf{0.7743} \\
F1 Score & 0.7060 & 0.7296 & \textbf{0.7944} \\
\bottomrule
\end{tabular}
\end{table}
}



\subsection{Analysis of results}

\begin{figure}[!h]
    \centering
    \includegraphics[width=0.95\linewidth]{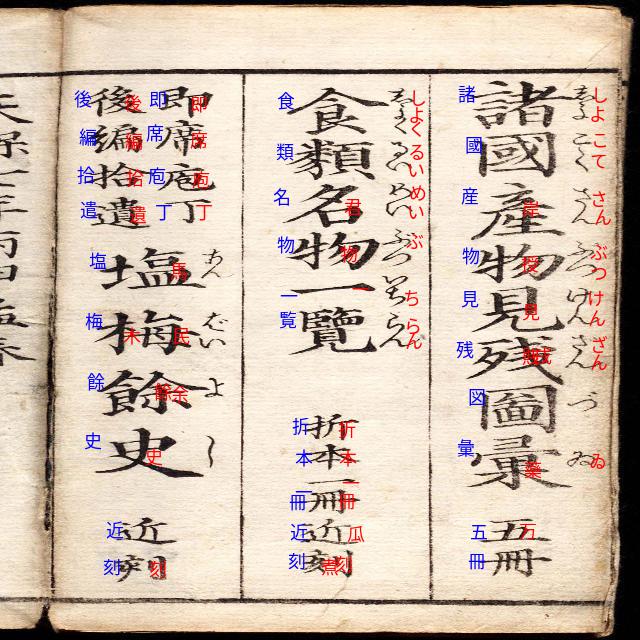}
    \caption{An example of large font sizes in the title page of a book, which causes our model to struggle (ground truth in blue, predictions in red).  The model struggles with these pages for three reasons.  First it struggles to detect and correctly recognize the large characters.  A second and more surprising issue is that the annotations (see Section~\ref{sec:challenges}) are smaller than the adjacent text, and as a result annotations on the title page are the same absolute size as the main text on a normal page.  Our model incorrectly classifies these annotations on a title page as main text.  Finally, these title pages appear at most once per book, and supervised learning often struggles when the amount of labeled data is small.  }
    \label{fig:bigfont}
\end{figure}

\begin{figure}
    \centering
    \includegraphics[width=0.4\linewidth]{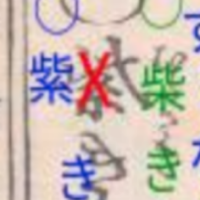}
    \includegraphics[width=0.4\linewidth]{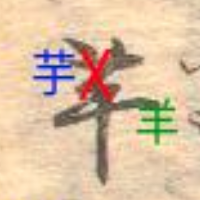}
    \caption{Two examples of the model predicting Kanji incorrectly (ground truth is blue, model's prediction is green), but with visual elements in common with the correct character.  }
    \label{fig:wrong}
\end{figure}

We have identified areas where our model still struggles: 
\begin{itemize}
    \item Difficulty in correctly recognizing characters which are very large in size.  We conjecture that this is due to these characters being relatively rare as seen in Figure~\ref{fig:bigfont}.  
    \item When predicting Kanji, sometimes a character is predicted which shares significant visual similarity with the correct character but is slightly different.  We give two examples of this in Figure~\ref{fig:wrong}.  
\end{itemize}

\section{Future Work}

The amount of unlabeled data which is available vastly exceeds the amount of labeled data, especially when one only considers the labeled data which is public and easily accessible (the dataset we used is public and has 27 labeled books). The Pre-Modern Japanese Text dataset, also from the same institution, contains 3126 unlabeled books), so this task could be an excellent setting for using semi-supervised learning.  This has been successfully demonstrated in a variety of settings.  \cite{dumoulin2016adversarially} demonstrated successful semi-supervised learning using latent variables inferred from generative models.  Strong results on semi-supervised learning have been demonstrated \citep{verma2019ict,beckham2019adversarial} using variants of the mixup algorithm, which we already use as a regularizer for supervised learning (Section~\ref{sec:reg}).  

An additional area for future work would be to create better methods to improve generalization on Kanji characters, which is challenging due to the Kanji alphabet's large vocabulary size.  In KuroNet, we model $p(y_{ij} \vert c_{ij}, x)$ as a multinomial distribution at each position - thus the final weight matrix in the character classification output layer has separate parameters for each character.  If we could somehow group or identify related kanji characters, then we could share more parameters between them and perhaps generalize much better.  There is extensive work from the Machine Learning literature on ``Few Shot'' learning, where few examples are available for a particular class \citep{santoro2016one}.  

A further issue is that the earliest labeled book in our training set is from 1643 (Table~\ref{tb:results}).  How KuroNet performs on data from much earlier books (and if it would benefit from them being added to the training data) remains a completely open question that we hope future work will be able to explore.  

Our current dataset and model does not contain labels for the annotations placed between text columns (Section~\ref{sec:challenges}).  Thus our model is trained to ignore these annotations.  In the future it would be useful to produce a model which can read them.  

\section{Conclusion}

We have proposed and experimentally evaluated a model for recognizing pre-modern Japanese text.  We have shown that several aspects of these documents make this task challenging, including the large vocabulary size, long-range interactions, and complicated layouts.  We have proposed and experimentally validated KuroNet, a new approach which addresses these challenges by jointly reading and recognizing complete pages of text.   We have also identified several open challenges in this area to motivate future work.  In particular, our model only trains on labeled data and does not take advantage of the plentiful unlabeled data which is available for this task.  KuroNet gives high recognition accuracy and uses no pre-processing when making predictions, making it easy to apply to real world data.  As a result, we hope that this will be an initial step towards making pre-modern Japanese books more accessible to the general public and preserving the cultural heritage of the Japanese people.




\bibliography{mybib}

\end{document}